\documentclass{article}

\usepackage[final]{nips_2018}

\usepackage[utf8]{inputenc} %
\usepackage[T1]{fontenc}    %
\usepackage{url}            %
\usepackage{booktabs}       %
\usepackage{amsfonts}       %
\usepackage{nicefrac}       %
\usepackage{microtype}      %

\usepackage{graphicx}
\graphicspath{{images/}}

\usepackage{color}

\definecolor{link_color}{RGB}{0,128,255}
\usepackage[%
colorlinks=true,allcolors=link_color,pageanchor=true,%
plainpages=false,pdfpagelabels,bookmarks,bookmarksnumbered,%
]{hyperref}

\usepackage{amsmath}
\usepackage{amsthm}
\usepackage{arydshln}
\usepackage{bm}
\usepackage{subcaption}

\usepackage[nameinlink]{cleveref}

\usepackage{algorithm}
\usepackage{algorithmicx}
\usepackage{algpseudocode}
\algnewcommand{\LeftComment}[1]{\Statex \(\triangleright\) #1}

\newcounter{module}
\makeatletter
\newenvironment{module}[1][htb]{%
  \let\c@algorithm\c@module
    \renewcommand{\ALG@name}{Module}%
   \begin{algorithm}[#1]%
  }{\end{algorithm}}
\makeatother
\crefname{module}{Module}{Modules}

\usepackage{tikz}
\usetikzlibrary{positioning}

\newcommand{\nwc}{\newcommand}
\nwc{\as}{\textrm{a.s.}}
\nwc{\defas}{:=}

\nwc{\dist}{\ \sim\ }
\nwc{\distiid}{\stackrel{\mathrm{iid}}{\sim}}

\DeclareMathOperator*{\argmin}{argmin}

\DeclareMathOperator*{\subjectto}{subject\;to}

\newcommand{\E}{\mathbb{E}}

\newcommand{\LL}{\mathcal{L}}

\newcommand{\xinit}{x_{\rm init}}
\newcommand{\uinit}{u_{\rm init}}
\newcommand{\LQR}{\ensuremath{\mathrm{LQR}}}
\newcommand{\MPC}{\ensuremath{\mathrm{MPC}}}

\newcommand{\RR}{\mathbb{R}}
\newcommand{\ZZ}{\mathbb{Z}}

\usepackage{accents}

\usepackage{wrapfig}

\title{\Large Differentiable MPC for End-to-end Planning and Control}

\author{
  Brandon Amos$^1$
  \quad
  Ivan Dario Jimenez Rodriguez$^2$
  \quad
  Jacob Sacks$^2$ \\
  \textbf{
  Byron Boots$^2$
  \quad
  J.~Zico Kolter$^{13}$} \\
  $^1$Carnegie Mellon University
  \qquad
  $^2$Georgia Tech
  \qquad
  $^3$Bosch Center for AI
}

\begin{document}

\maketitle

\begin{abstract}
  We present foundations for using
  Model Predictive Control (MPC) as a differentiable policy
  class for reinforcement learning in continuous
  state and action spaces.
  This provides one way of leveraging and combining the advantages
  of model-free and model-based approaches.
  Specifically, we differentiate through MPC by using the
  KKT conditions of the convex approximation at a fixed
  point of the controller.
  Using this strategy, we are able to learn the cost and
  dynamics of a controller via end-to-end learning.
  Our experiments focus on imitation learning in the pendulum
  and cartpole domains, where we learn the cost and dynamics
  terms of an MPC policy class.
  We show that our MPC policies are significantly more data-efficient
  than a generic neural network and that our method is
  superior to traditional system identification
  in a setting where the expert is unrealizable.
\end{abstract}

\section{Introduction}
Model-free reinforcement learning has achieved state-of-the-art 
results in many challenging domains.
However, these methods learn black-box control policies and typically 
suffer from poor sample complexity and generalization.
Alternatively, model-based approaches seek to model the environment the
agent is interacting in.
Many model-based approaches utilize Model Predictive Control
(MPC) to perform complex control tasks
\citep{gonzalez2011robust,lenz2015deepmpc,liniger2014rccar,kamel2015multicopter, erez2012mpc,alexis2011quadrotor,bouffard2012lbmpc,neunert2016slqmpc}.
MPC leverages a predictive model of the controlled system and solves 
an optimization problem online in a receding horizon fashion to 
produce a sequence of control actions.
Usually the first control action is applied to the system,
after which the optimization problem is solved again for the next time step.

Formally, MPC requires that at each time step we solve the optimization problem:
\begin{equation}
    \argmin_{x_{1:T} \in \mathcal{X},u_{1:T}\in \mathcal{U}} \;\; \sum_{t=1}^T  C_t(x_t, u_t) 
    \;\; \subjectto \;\; x_{t+1} = f(x_t, u_t), \;\; x_1 = \xinit,
  \label{eq:mpc}
\end{equation}
where $x_t, u_t$ are the state and control at time $t$, $\mathcal{X}$ and
$\mathcal{U}$ are constraints on valid states and controls, $C_t : \mathcal{X}
\times \mathcal{U} \rightarrow \mathbb{R}$ is a (potentially time-varying) cost
function, $f : \mathcal{X} 
\times \mathcal{U} \rightarrow \mathcal{X}$ is a dynamics model, and
$\xinit$ is the initial state of the system.  
The optimization problem in \cref{eq:mpc} can be efficiently solved 
in many ways, for example with the finite-horizon iterative Linear
Quadratic Regulator (iLQR) algorithm \citep{li2004ilqr}. Although these
techniques are widely used in control domains, much work in
deep reinforcement learning or imitation learning opts instead to use a much
simpler policy class such as a linear function or neural network.
The advantages of these policy classes is that they are
differentiable and the loss can be directly optimized with respect to them
while it is typically not possible to do full end-to-end
learning with model-based approaches.

\begin{figure}[t]
  \centering
  \includegraphics[width=0.8\textwidth]{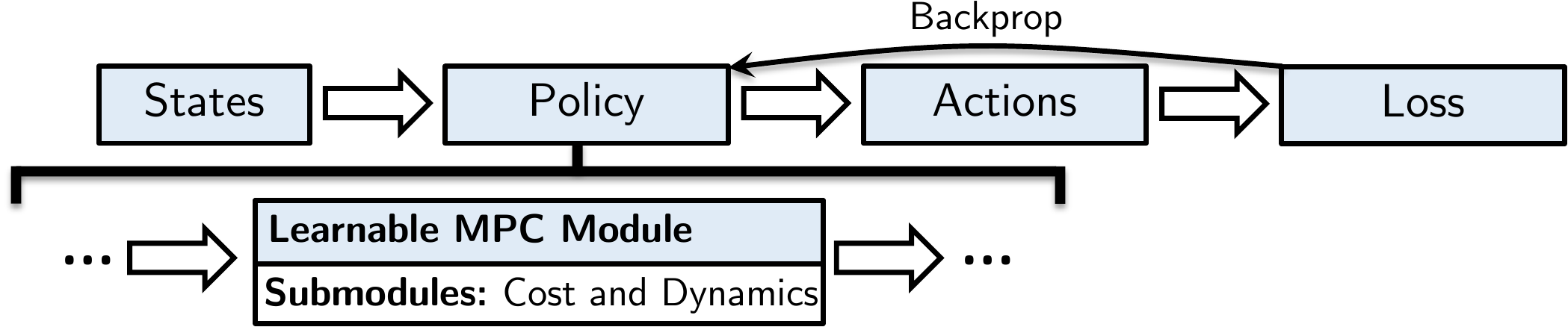}
  \caption{
    \textbf{Illustration of our contribution:} A learnable MPC module that
    can be integrated into a larger end-to-end reinforcement learning
    pipeline.
    Our method allows the controller to be updated with gradient
    information directly from the task loss.
  }
  \label{fig:overview}
\end{figure}

In this paper, we consider the task of learning MPC-based policies in an
end-to-end fashion, illustrated in \cref{fig:overview}.
That is, we treat MPC as a generic policy
class $u = \pi(\xinit; C, f)$ parameterized by some representations of
the cost $C$ and dynamics model $f$.
By differentiating \emph{through} the optimization
problem, we can learn the costs and dynamics model to perform
a desired task. This is in contrast to regressing on collected dynamics or
trajectory rollout data and learning each component in isolation, and comes with
the typical advantages of end-to-end learning (the ability to train directly
based upon the task loss of interest, the ability to ``specialize'' parameter
for a given task, etc).

Still, efficiently differentiating through a complex 
policy class like MPC is challenging.
Previous work with similar aims has either simply unrolled and differentiated 
through a simple optimization procedure \citep{tamar2017learning} or 
has considered generic optimization solvers that do not scale to the 
size of MPC problems \citep{amos2017optnet}.
This paper makes the following two contributions to this space.
First, we provide an efficient method for \emph{analytically} differentiating 
through an iterative non-convex optimization procedure based upon a 
box-constrained iterative LQR solver \citep{tassa2014control}; in particular, 
we show that the analytical derivative can be computed using 
\emph{one additional} backward pass of a modified iterative LQR solver.  
Second, we empirically show that in imitation learning scenarios
we can recover the \emph{cost} and \emph{dynamics} from an MPC
expert with a loss based only on the actions (and not states).
In one notable experiment, we show that directly optimizing the
imitation loss results in better performance than vanilla
system identification.

\section{Background and Related Work}
\textbf{Pure model-free techniques for policy search} have
demonstrated promising results in many domains by learning
\emph{reactive polices} which directly map observations to actions
\citep{mnih2013playing,oh2016minecraft,gu2016continuous,lillicrap2015continuous,
schulman2015trust,schulman2016trpogae,gu2017qprop}.
Despite their success, model-free methods have many drawbacks and limitations, 
including a lack of interpretability, poor generalization, and a 
high sample complexity.
\textbf{Model-based methods} are known to be more sample-efficient
than their model-free 
counterparts.
These methods generally rely on learning a dynamics model directly from 
interactions with the real system and then integrate the learned model into the
control policy
\citep{schneider1997exploiting,abbeel2006using,deisenroth2011pilco,heess2015learning,
boedecker2014sparsegps}.
More recent approaches use a deep network to learn low-dimensional latent state
representations and associated dynamics models in this learned representation. 
They then apply standard trajectory optimization methods
on these learned embeddings
\citep{lenz2015deepmpc, watter2015embed, levine2016end}.
However, these methods still require a manually specified and hand-tuned
cost function, which can become even more difficult in a latent representation.
Moreover, there is no guarantee that the learned dynamics model
can accurately capture portions of the state space relevant for the task at hand.

To leverage the benefits of both approaches, there has been significant
interest in \textbf{combining the model-based and model-free paradigms.}
In particular, much attention has been dedicated to utilizing
model-based priors to accelerate the model-free learning process.
For instance, synthetic training data can be generated by model-based control 
algorithms to guide the policy search or prime a 
model-free policy
\citep{sutton1990integrated,theodorou2010generalized,levine2014learning,
gu2016continuous,venkatraman2016improved,levine2016end,chebotar2017combining, 
nagabandi2017mbmf, liting2017driving}.
\citep{bansal2017mbmf} learns a controller and then distills it to
a neural network policy which is then fine-tuned with model-free
policy learning.
However, this line of work usually keeps the model separate from the
learned policy.

Alternatively, the policy can include an \textbf{explicit planning module}
which \emph{leverages learned models} of the system or environment,
both of which are learned through model-free techniques.
For example, the classic Dyna-Q algorithm
\citep{sutton1990integrated} simultaneously learns a model of the environment
and uses it to plan.
More recent work has explored incorporating such structure into deep
networks and learning the policies in an end-to-end fashion.
\cite{tamar2016value} uses a recurrent network to predict the value function by 
approximating the value iteration algorithm with convolutional layers.
\cite{karkus2017qmdp} connects a dynamics model to a planning
algorithm and formulates the policy as a structured recurrent network.
\cite{silver2016predictron} and \cite{oh2017value} perform multiple rollouts
using an abstract dynamics model to predict the value function.
A similar approach is taken by \cite{weber2017imagination} but directly 
predicts the next action and reward from rollouts of an explicit environment model.
\cite{farquhar2017treeqn} extends model-free approaches, such as 
DQN \citep{mnih2015human} and A3C \citep{mnih2016asynchronous}, by planning 
with a tree-structured neural network to predict the cost-to-go.
While these approaches have demonstrated impressive results in
discrete state and action spaces, they are not applicable to
continuous control problems.

To tackle continuous state and action spaces, \cite{pascanu2017learning}
propose a neural architecture which uses an abstract environmental 
model to plan and is trained directly from an external task loss.
\cite{pong2018temporal} learn goal-conditioned value functions and use them
to plan single or multiple steps of actions in an MPC fashion.
Similarly, \cite{pathak2018zero} train a goal-conditioned policy to perform
rollouts in an abstract feature space but ground the policy with a loss term
which corresponds to true dynamics data.
The aforementioned approaches can be interpreted as a distilled optimal controller
which does not separate components for the cost and dynamics.
Taking this analogy further, another strategy is to differentiate through an
optimal control algorithm itself.
\cite{okada2017path} and \cite{pereira2018pinets} present a way
to differentiate through path integral optimal control
\citep{williams2016aggressive,williams2017model}
and learn a planning policy end-to-end.
\cite{srinivas2018universal} shows how to embed
differentiable planning (unrolled gradient descent over actions) within
a goal-directed policy.
In a similar vein, \cite{tamar2017learning} differentiates 
through an iterative LQR (iLQR) solver
\citep{li2004ilqr,xie2017ddp,tassa2014control} 
to learn a cost-shaping term offline.
This shaping term enables a shorter horizon controller to approximate the
behavior of a solver with a longer horizon to save computation during runtime.

\textbf{Contributions of our paper.}  
All of these methods require differentiating through planning procedures
by explicitly ``unrolling'' the optimization algorithm itself.
While this is a reasonable strategy, it is both memory- and
computationally-expensive and challenging when unrolling
through many iterations because the time- and space-complexity
of the backward pass grows linearly with the forward pass.
In contrast, we address this issue by showing
how to \emph{analytically} differentiate through the fixed point of a
nonlinear MPC solver.
Specifically, we compute the derivatives of an iLQR solver with a
\emph{single} LQR step in the backward pass. This makes the learning process
more computationally tractable while still allowing us to plan in continuous
state and action spaces. Unlike model-free approaches, explicit cost and dynamics
components can be extracted and analyzed on their own. Moreover, in contrast to
pure model-based approaches, the dynamics model and cost function can be learned
entirely end-to-end.

\section{Differentiable LQR}
\label{sec:diff-lqr}

Discrete-time finite-horizon LQR is a well-studied control method
that optimizes a convex quadratic objective function
with respect to affine state-transition dynamics
from an initial system state $\xinit$.
Specifically, LQR finds the optimal nominal trajectory
$\tau_{1:T}^\star = \{x_t, u_t\}_{1:T}$
by solving the optimization problem
\begin{equation}
  \label{eq:lqr}
    \tau^{\star}_{1:T} = \argmin_{\tau_{1:T}}\;\; 
    \sum_{t=1}^T \frac{1}{2} \tau_t^\top  C_t \tau_t + c_t^\top  \tau_t \;\;
    \subjectto\;\;
    x_1 = \xinit,\
    x_{t+1} = F_t\tau_t + f_t.
\end{equation}
From a policy learning perspective, this can be interpreted as a 
module with unknown parameters $\theta=\{C, c, F, f\}$, which can 
be integrated into a larger end-to-end learning system.
The learning process involves taking derivatives of some loss function 
$\ell$, which are then used to update the parameters.
Instead of directly computing each of the individual gradients, we present 
an efficient way of computing the derivatives of the loss function with 
respect to the parameters
\begin{equation}
  \frac{\partial\ell}{\partial\theta} =
  \frac{\partial\ell}{\partial\tau_{1:T}^\star}
  \frac{\partial\tau_{1:T}^\star}{\partial\theta}.
\end{equation}
\newpage
By interpreting LQR from an optimization perspective
\citep{boyd2008lqr}, we associate dual variables
$\lambda_{0:T-1}$ with the state constraints, where
$\lambda_0$ is associated with the $x_1=\xinit$ constraint
and $\lambda_{1:T-1}$ are associated with the remaining ones,
respectively.
The Lagrangian of the optimization problem is then
\begin{align}
  \label{eq:lagrangian}
  \mathcal{L}(\tau, \lambda) =
    \sum_{t=1}^T \left(\frac{1}{2}\tau_t^\top  C_t \tau_t + c_t^\top \tau_t\right) +
    \sum_{t=0}^{T-1} \lambda_t^\top  (F_t\tau_t + f_t - x_{t+1}),
\end{align}
where the initial constraint $x_1=\xinit$ is represented by
setting $F_0=0$ and $f_0=\xinit$.
Differentiating \cref{eq:lagrangian} with respect to
$\tau_t^\star$ yields
\begin{align}
  \label{eq:lqr_lagrangian_diff}
  \nabla_{\tau_t}\mathcal{L}(\tau^\star, \lambda^\star) =
     C_t \tau_t^\star + c_t + F_t^\top \lambda_t^\star -
  \begin{bmatrix}
    \lambda_{t-1}^\star \\
    0
  \end{bmatrix}
  = 0,
\end{align}
where the $\lambda_{t-1}^\star$ term comes from the $x_{t+1}$
term in \cref{eq:lagrangian}.

\begin{module}[t]
\footnotesize
\caption{Differentiable LQR \hfill
  \textit{\footnotesize
    (The LQR algorithm is defined in \cref{sec:mpc-algs})}}
\label[module]{alg:differentiable-lqr}
\textbf{Input:} Initial state $\xinit$ \\
\textbf{Parameters:} $\theta=\{C, c, F, f\}$ \\~\\
\textbf{Forward Pass:}
\begin{algorithmic}[1]
  \State $\tau_{1:T}^\star = \LQR_T(\xinit; C, c, F, f)$
  \Comment{Solve \eqref{eq:lqr}}
  \State Compute $\lambda_{1:T}^\star$ with \eqref{eq:lqr_adjoint_lambdas}
\end{algorithmic}~\\
\textbf{Backward Pass:}
\begin{algorithmic}[1]
  \State $d_{\tau_{1:T}}^\star= \LQR_T(0; C, \nabla_{\tau^\star} \ell, F, 0)$
  \Comment{Solve \eqref{eq:lqr_diff_system},
    ideally reusing the factorizations from the forward pass
  }
  \State Compute $d_{\lambda_{1:T}}^\star$ with \eqref{eq:lqr_adjoint_lambdas}
  \State Compute the derivatives of $\ell$ with respect to
  $C$, $c$, $F$, $f$, and $\xinit$ with
  \eqref{eq:lqr_derivatives}
\end{algorithmic}
\end{module}

Thus, the normal approach to solving LQR problems with dynamic Riccati recursion
can be viewed as an efficient way of solving the KKT system
\begin{align}
  \label{eq:lqr_kkt_system}
  \overbrace{
  \begin{array}{c}
    \begin{array}{cccccc}
      &\tau_t \quad &\lambda_t&& \tau_{t+1}& \quad \lambda_{t+1}
    \end{array} \\
    \left[\begin{array}{c;{2pt/2pt}cc;{2pt/2pt}cc;{2pt/2pt}c}
            \ddots &&&&&\\ \hdashline[2pt/2pt]
                   & C_t & F_t^\top  & &  & \\
                   & F_t &  & [-I \quad 0]  &  & \\ \hdashline[2pt/2pt]
                   & &
                       \begin{bmatrix}
                         -I \\
                         0
                       \end{bmatrix}
                   & C_{t+1} & F_{t+1}^\top  & \\
                   & & & F_{t+1} &  & \\ \hdashline[2pt/2pt]
                   & & & &  & \ddots \\
          \end{array}
    \right]
  \end{array}
  }^{K}
  \begin{bmatrix}
    \vdots \\
    \tau_t^\star \\
    \lambda_t^\star \\    
    \tau_{t+1}^\star \\
    \lambda_{t+1}^\star \\        
    \vdots \\
  \end{bmatrix}
  = -
  \begin{bmatrix}
    \vdots \\
    c_t\\
    f_t \\
    c_{t+1} \\
    f_{t+1} \\
    \vdots \\
  \end{bmatrix}.
\end{align}

Given an optimal nominal trajectory $\tau_{1:T}^\star$,
\cref{eq:lqr_lagrangian_diff} shows how to compute the
optimal dual variables $\lambda$ with the backward recursion
\begin{align}
  \label{eq:lqr_adjoint_lambdas}
  \begin{split}
    \lambda_T^\star &= C_{T,x}\tau_{T}^\star + c_{T,x} \qquad
    \lambda_t^\star = F_{t,x}^\top  \lambda_{t+1}^\star + C_{t,x}\tau_t^\star + c_{t,x},
\end{split}
\end{align}
where $C_{t,x}$, $c_{t,x}$, and $F_{t,x}$ are the first
block-rows of $C_t$, $c_t$, and $F_t$, respectively.
Now that we have the optimal trajectory and dual variables, we can compute 
the gradients of the loss with respect to the parameters.
Since LQR is a constrained convex quadratic $\argmin$, the derivatives
of the loss with respect to the LQR parameters can be obtained by
implicitly differentiating the KKT conditions.
Applying the approach from Section~3 of \cite{amos2017optnet},
the derivatives are
\begin{align}
  \label{eq:lqr_derivatives}
  \begin{aligned}
    \nabla_{C_t} \ell &=
    \frac{1}{2}\left(d_{\tau_t}^\star \otimes \tau_t^\star +
      \tau_t^\star \otimes d_{\tau_t}^\star \right)\qquad &
    \; \nabla_{c_t} \ell &= d_{\tau_t}^\star \qquad &
      \nabla_{\xinit} \ell &= d_{\lambda_0}^\star \\
    \nabla_{F_t} \ell &=
    d_{\lambda_{t+1}}^\star \otimes \tau_t^\star +
    \lambda_{t+1}^\star \otimes d_{\tau_t}^\star\qquad  & 
    \; \nabla_{f_t} \ell &= d_{\lambda_t}^\star \\
\end{aligned}
\end{align}
where $\otimes$ is the outer product operator,
and $d_\tau^\star$ and $d_\lambda^\star$ are obtained by solving
the linear system
\begin{align}
  \label{eq:lqr_diff_system}
  K
  \begin{bmatrix}
    \vdots\\
    d_{\tau_t}^\star \\
    d_{\lambda_t}^\star \\
    \vdots
  \end{bmatrix}
  = -
  \begin{bmatrix}
    \vdots \\
    \nabla_{\tau_t^\star} \ell \\
    0 \\
    \vdots
  \end{bmatrix}.
\end{align}
We observe that \cref{eq:lqr_diff_system} is of the same form
as the linear system in \cref{eq:lqr_kkt_system} for the LQR problem.
Therefore, we can leverage this insight and solve \cref{eq:lqr_diff_system} 
efficiently by solving another LQR problem that replaces
$c_t$ with $\nabla_{\tau_t^\star} \ell$ and $f_t$ with 0.
Moreover, this approach enables us to re-use the factorization of $K$ from the
forward pass instead of recomputing.
\cref{alg:differentiable-lqr} summarizes the forward and backward passes for
a differentiable LQR module.

\section{Differentiable MPC}

While LQR is a powerful tool, it does not cover
realistic control problems with non-linear
dynamics and cost.
Furthermore, most control problems have natural bounds on the
control space that can often be expressed as box constraints.
These highly non-convex problems, which we will refer to as
model predictive control (MPC), are well-studied in the
control literature and can be expressed in the general form
\begin{equation}
  \label{eq:ilqr}
    \tau_{1:T}^{\star} = \argmin_{\tau_{1:T}}\;\; \sum_tC_{\theta,t}(\tau_t) \;\;
    \subjectto\;\;
    x_1 = \xinit, \
    x_{t+1} = f_\theta(\tau_t), \
    \underline{u} \leq u \leq \overline{u},
\end{equation}

where the non-convex cost function $C_\theta$ and non-convex
dynamics function $f_\theta$ are (potentially) parameterized
by some $\theta$.
We note that more generic constraints on the control
and state space can be represented as penalties and barriers
in the cost function.
The standard way of solving the control problem
in \cref{eq:ilqr} is by iteratively forming and
optimizing a convex approximation
\begin{equation}
  \label{eq:ilqr_convex}
    \tau_{1:T}^i = \argmin_{\tau_{1:T}}\;\; \sum_t \tilde C_{\theta,t}^i(\tau_t) \;\;
    \subjectto\;\;
    x_1 = \xinit,\ 
    x_{t+1} = \tilde f_\theta^i(\tau_t),\
    \underline{u} \leq u \leq \overline{u},
\end{equation}
where we have defined the second-order Taylor approximation of the
cost around $\tau^i$ as
\begin{equation}
  \label{eq:ilqr-cost-taylor}
  \tilde C_{\theta,t}^i = C_{\theta,t}(\tau_t^i) + (p_t^i)^\top (\tau_t-\tau_t^i) +
  \frac{1}{2} (\tau_t-\tau_t^i)^\top  H_t^i(\tau_t-\tau_t^i)
\end{equation}
with $p_t^i=\nabla_{\tau_t^i} C_{\theta,t}$ and
$H_t^i=\nabla_{\tau_t^i}^2 C_{\theta,t}$.
We also have a first-order Taylor approximation of the dynamics around $\tau^i$
as
\begin{equation}
  \label{eq:ilqr-dynamics-taylor}
  \tilde f_{\theta,t}^i(\tau_t) = f_{\theta,t}(\tau_t^i) + F_t^i(\tau_t-\tau_t^i)
\end{equation}
with $F_t^i=\nabla_{\tau_t^i} f_{\theta,t}$.
In practice, a fixed point of \cref{eq:ilqr_convex} is often
reached, especially when the dynamics are smooth.
As such, differentiating the non-convex problem \cref{eq:ilqr}
can be done exactly by using the final
convex approximation.
Without the box constraints, the fixed point in
\cref{eq:ilqr_convex} could be differentiated
with LQR as we show in \cref{sec:diff-lqr}.
In the next section, we will show how to extend this to the case where we have box constraints on the controls as well.

\subsection{Differentiating Box-Constrained QPs}
First, we consider how to differentiate a more generic
box-constrained convex QP of the form
\begin{equation}
  \label{eq:box_qp}
  x^\star =  \argmin_x \;\; \frac{1}{2} x^\top  Q x + p^\top  x \;\;
  \subjectto\;\; Ax = b, \
  \underline{x} \leq x \leq \overline{x}.
\end{equation}
Given active inequality constraints at the solution
in the form $\tilde Gx = \tilde h$,
this problem turns into an equality-constrained
optimization problem with the solution given
by the linear system
\begin{align}
  \label{eq:qp_active_solve}
  \begin{bmatrix}
    Q & A^\top  & \tilde G^\top  \\
    A & 0 & 0 \\
    \tilde G & 0 & 0 \\
  \end{bmatrix}
  \begin{bmatrix}
    x^\star \\ \lambda^\star \\ \tilde \nu^\star
  \end{bmatrix}
  = -
  \begin{bmatrix}
      p \\ b \\ \tilde h
  \end{bmatrix}
\end{align}
With some loss function $\ell$ that depends on $x^\star$,
we can use the approach in \cite{amos2017optnet}
to obtain the derivatives of $\ell$ with respect
to $Q$, $p$, $A$, and $b$ as
\begin{equation}
  \label{eq:qp_derivatives}
    \nabla_Q \ell =
      \frac{1}{2}(d_x^\star \otimes x^\star + x^\star \otimes d_x^\star) \qquad
    \nabla_p \ell = d_x^\star \qquad
    \nabla_A \ell =
      d_\lambda^\star \otimes x^\star + \lambda^\star \otimes d_x^\star \qquad
    \nabla_b \ell = -d_\lambda^\star \\
\end{equation}
where $d_x^\star$ and $d_\lambda^\star$ are obtained by solving the linear system
\begin{align}
  \label{eq:qp_diff_system}
  \begin{bmatrix}
    Q & A^\top  & \tilde G^\top  \\
    A & 0 & 0 \\
    \tilde G & 0 & 0 \\
  \end{bmatrix}
  \begin{bmatrix}
    d_x^\star \\ d_\lambda^\star \\ d_{\tilde \nu}^\star
  \end{bmatrix}
  = -
  \begin{bmatrix}
      \nabla_{x^\star} \ell \\ 0 \\ 0
  \end{bmatrix}
\end{align}
The constraint $\tilde G d_x^\star = 0$ is equivalent to
the constraint $d_{x_i}^\star=0$ if
$x_i^\star \in \{\underline{x}_i, \overline{x}_i\}$.
Thus solving the system in \cref{eq:qp_diff_system} is
equivalent to solving the optimization problem
\begin{align}
  \label{eq:box_qp_diff}
  d_x^\star =  \argmin_{d_x} \;\; & \frac{1}{2} d_x^\top  Q d_x + (\nabla_{x^\star} \ell)^\top  d_x \;\;
  \subjectto\;\; A d_x = 0,\ 
  d_{x_i} = 0\;\; \text{if}\;\; x_i^\star \in\{\underline{x}_i, \overline{x}_i\}
\end{align}

\subsection{Differentiating MPC with Box Constraints}
\begin{module}[t]
\footnotesize
\caption{Differentiable MPC \hfill
  \textit{\footnotesize
    (The MPC algorithm is defined in \cref{sec:mpc-algs})}}
\label[module]{alg:differentiable-mpc}
\textbf{Given:} Initial state $\xinit$ and initial control sequence $\uinit$ \\
\textbf{Parameters:} $\theta$ of the objective $C_\theta(\tau)$ and
  dynamics $f_\theta(\tau)$ \\~\\
\textbf{Forward Pass:}
\begin{algorithmic}[1]
  \State $\tau_{1:T}^\star = \MPC_{T,\underline{u},\overline{u}}(\xinit, \uinit; C_\theta, F_\theta)$
  \Comment{Solve \cref{eq:ilqr}}
  \State \emph {The solver should reach the fixed
    point in \eqref{eq:ilqr_convex} to obtain approximations to the
    cost $H^n_\theta$ and dynamics $F^n_\theta$}
  \State Compute $\lambda_{1:T}^\star$ with \eqref{eq:lqr_adjoint_lambdas}
\end{algorithmic}~\\
\textbf{Backward Pass:}
\begin{algorithmic}[1]
  \State $\tilde F^n_\theta$ is $F^n_\theta$ with the rows corresponding
  to the tight control constraints zeroed

  \State $d_{\tau_{1:T}}^\star= \LQR_T(0; H_\theta^n, \nabla_{\tau^\star} \ell,
    \tilde F^n_\theta, 0)$
  \Comment{Solve \eqref{eq:mpc_diff},
    ideally reusing the factorizations from the forward pass
  }
  \State Compute $d_{\lambda_{1:T}}^\star$ with \eqref{eq:lqr_adjoint_lambdas}
  \State Differentiate $\ell$ with respect to
  the approximations $H^n_\theta$ and $F^n_\theta$
  with \eqref{eq:lqr_derivatives}
  \State Differentiate these approximations
  with respect to $\theta$ and use the chain rule
  to obtain $\partial \ell/\partial \theta$
\end{algorithmic}
\end{module}

At a fixed point, we can use
\cref{eq:qp_derivatives} to compute the derivatives of the MPC problem,
where $d_\tau^\star$ and $d_\lambda^\star$ are found by
solving the linear system in
\cref{eq:lqr_diff_system}
with the additional constraint that
$d_{u_{t,i}}=0$ if $u_{t,i}^\star\in\{\underline{u}_{t,i}, \overline{u}_{t,i}\}$.
Solving this system can be equivalently written as a
zero-constrained LQR problem of the form
\begin{align}
  \label{eq:mpc_diff}
  \begin{split}
    d_{\tau_{1:T}}^{\star} = \argmin_{d_{\tau_{1:T}}} \;\; &
    \sum_{t} \frac{1}{2} d_{\tau_t}^\top  H_t^n d_{\tau_t} +
      (\nabla_{\tau_t^\star} \ell)^\top  d_{\tau_t} \\
    \subjectto\;\;
    & d_{x_1} = 0,\ 
    d_{x_{t+1}} = F_t^n d_{\tau_t},\ 
    d_{u_{t,i}} = 0\;\; \text{if}\;\; u_i^\star
      \in\{\underline{u}_{t,i}, \overline{u}_{t,i}\} \\
  \end{split}
\end{align}
where $n$ is the iteration that \cref{eq:ilqr_convex}
reaches a fixed point, and $H^n$ and $F^n$ are the
corresponding approximations to the objective and
dynamics defined earlier.
\cref{alg:differentiable-mpc} summarizes the proposed
differentiable MPC module.
To solve the MPC problem in \cref{eq:ilqr} and reach the fixed point in
\cref{eq:ilqr_convex}, we use the box-DDP heuristic \citep{tassa2014control}.
For the zero-constrained LQR problem in \cref{eq:mpc_diff}
to compute the derivatives, we use an LQR solver that zeros
the appropriate controls.

\subsection{Drawbacks of Our Approach}
Sometimes the controller does not run for long enough to reach a
fixed point of \cref{eq:ilqr_convex}, or a fixed
point doesn't exist, which often happens when
using neural networks to approximate the dynamics.
When this happens, \cref{eq:mpc_diff} cannot be used to differentiate
through the controller, because it assumes a fixed point.
Differentiating through the final iLQR 
iterate that's not a fixed point will usually
give the wrong gradients.
Treating the iLQR procedure as a compute graph and differentiating through
the unrolled operations is a reasonable alternative in this scenario that
obtains surrogate gradients to the control problem.
However, as we empirically show
in \cref{sec:timing},
the backward pass of this method
scales linearly with the
number of iLQR iterations used in the forward.
Instead, fixed-point differentiation is constant time
and only requires a single LQR solve.

\section{Experimental Results}
In this section, we present several results that highlight the
performance and capabilities of differentiable MPC in comparison
to neural network policies and vanilla system identification (SysId).
We show
1) superior runtime performance compared to
an unrolled solver,
2) the ability of our method to recover the cost and dynamics
of a controller with imitation,
and
3) the benefit of directly optimizing the task loss
over vanilla SysId.

We have released our differentiable MPC solver as a standalone open
source package that is available at
\url{https://github.com/locuslab/mpc.pytorch}
and our experimental code for this paper is also openly available at
\url{https://github.com/locuslab/differentiable-mpc}.
Our experiments are implemented with PyTorch \citep{paszke2017automatic}.

\begin{figure}[t]
\centering
\begin{minipage}{.48\textwidth}
  \centering
  \includegraphics[width=\textwidth]{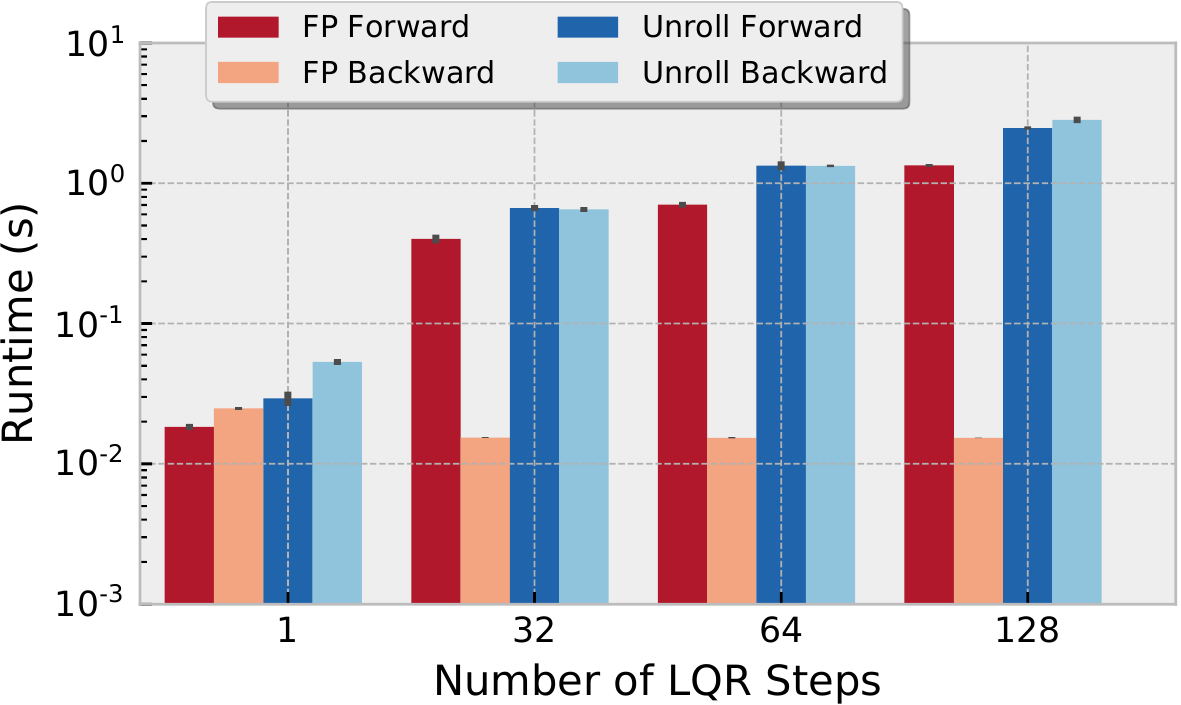}
  \captionof{figure}{
    Runtime comparison of fixed point differentiation (FP)
    to unrolling the iLQR solver (Unroll), averaged over 10 trials.}
  \label{fig:mpc_performance}
\end{minipage}\hfill
\begin{minipage}{.48\textwidth}
  \centering
  \vspace{6mm}
  \includegraphics[width=\textwidth]{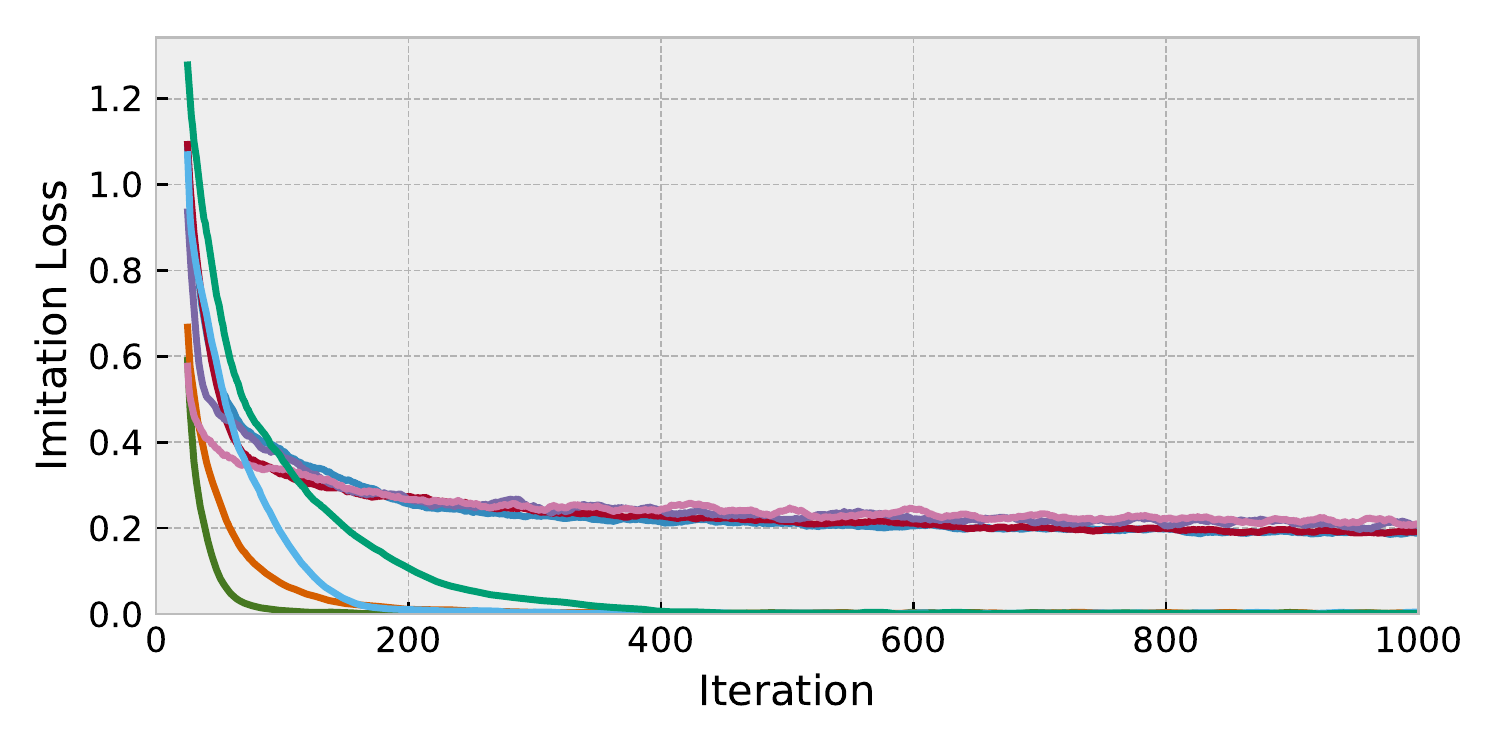}
  \includegraphics[width=\textwidth]{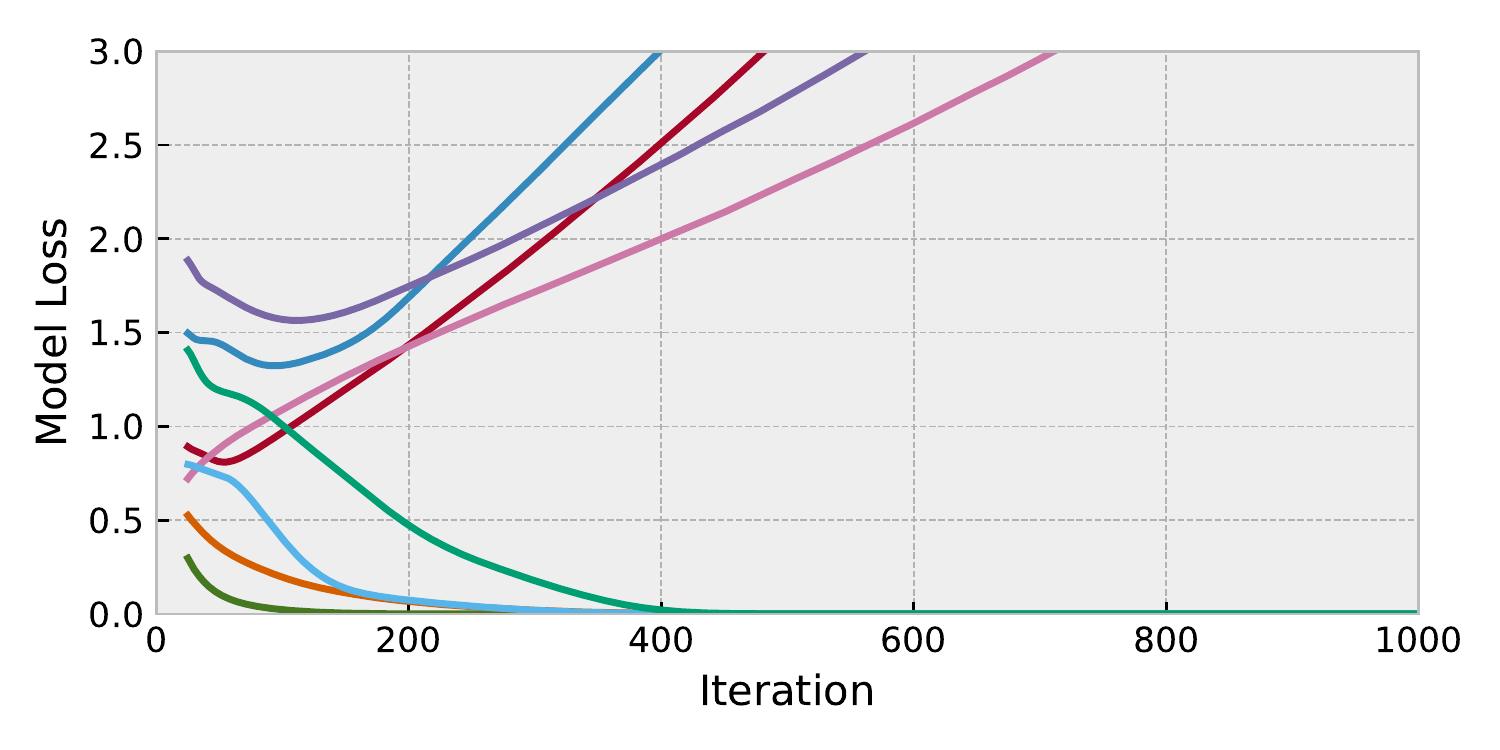}
  \captionof{figure}{
    Model and imitation losses for the LQR imitation learning experiments.
  }
  \label{fig:linear-model}
\end{minipage}
\end{figure}

\subsection{MPC Solver Performance}
\label{sec:timing}
\Cref{fig:mpc_performance} highlights the performance of our
differentiable MPC solver. We compare to an alternative
version where each box-constrained iLQR iteration is individually unrolled, and
gradients are computed by differentiating through the entire unrolled chain.
As illustrated in the figure, these unrolled operations incur a substantial
extra cost.
Our differentiable MPC solver 1) is slightly more computationally
efficient even in the forward pass, as it does not need to
create and maintain the backward pass variables; 2) is more memory
efficient in the forward pass for this same reason (by a factor of the
number of iLQR iterations);
and 3) is \emph{significantly} more efficient in the backward pass,
especially when a large number of iLQR iterations are needed. 
The backward pass is essentially free, as it can reuse all
the factorizations for the forward pass and does
not require multiple iterations.

\subsection{Imitation Learning: Linear-Dynamics Quadratic-Cost (LQR)}

In this section, we show results to validate the MPC solver and
gradient-based learning approach for an imitation learning problem.
The expert and learner are LQR controllers that share all information
except for the linear system dynamics $f(x_t, u_t) = Ax_t + Bu_t$.
The controllers have the same quadratic cost (the identity),
control bounds $[-1, 1]$, horizon (5 timesteps),
and 3-dimensional state and control spaces.
Though the dynamics can also be recovered by
fitting next-state transitions, we show that we can
alternatively use imitation learning to
recover the dynamics using only controls.

Given an initial state $x$, we can obtain nominal actions from the
controllers as $u_{1:T}(x; \theta)$, where $\theta=\{A, B\}$.
We randomly initialize the learner's dynamics with $\hat \theta$ and
minimize the \textbf{imitation loss}
$$\LL = \E_x\left[ ||\tau_{1:T}(x; \theta) - \tau_{1:T}(x; \hat\theta)||_2^2\right],.$$
We do learning by differentiating $\LL$ with
respect to $\hat \theta$ (using mini-batches with 32 examples) and
taking gradient steps with RMSprop \citep{tieleman2012lecture}.
\Cref{fig:linear-model} shows the model and imitation loss of
eight randomly sampled initial dynamics,
where the \textbf{model loss} is ${\rm MSE}(\theta, \hat\theta)$.
The model converges to the true parameters in half of the trials
and achieves a perfect imitation loss.
The other trials get stuck in a local minimum of the
imitation loss and causes the approximate model to significantly
diverge from the true model.
These faulty trials highlight that despite the LQR problem
being convex, the optimization problem of some loss function
w.r.t.~the controller's parameters is a (potentially difficult)
non-convex optimization problem that typically does not have
convergence guarantees.

\subsection{Imitation Learning: Non-Convex Continuous Control}
\label{il:ctrl}
We next demonstrate the ability of our method to do imitation
learning in the pendulum and cartpole benchmark domains.
Despite being simple tasks, they are relatively challenging for a
generic poicy to learn quickly in the imitation learning setting.
In our experiments we use MPC experts and learners
that produce a nominal action sequence
$u_{1:T}(x; \theta)$ where $\theta$ parameterizes the
model that's being optimized.
The goal of these experiments is to optimize the imitation loss
$\LL = \E_x\left[ ||u_{1:T}(x; \theta) - u_{1:T}(x; \hat\theta)||_2^2\right]$,
again which we can uniquely do using \emph{only} observed controls and
\emph{no} observations.
We consider the following methods:

\textbf{Baselines:}
\emph{nn} is an LSTM that takes the state $x$ as input and
predicts the nominal action sequence.
In this setting we optimize the imitation loss directly.
\emph{sysid} assumes the cost of the controller is known and
approximates the parameters of the dynamics by optimizing the
next-state transitions.

\textbf{Our Methods:}
\emph{mpc.dx}
assumes the cost of the controller is known and
approximates the parameters of the dynamics by directly
optimizing the imitation loss.
\emph{mpc.cost} assumes the dynamics of the controller
is known and approximates the cost by directly optimizing
the imitation loss.
\emph{mpc.cost.dx} approximates both the cost and parameters of the
dynamics of the controller by directly optimizing
the imitation loss.

In all settings that involve learning the dynamics
(\emph{sysid}, \emph{mpc.dx}, and \emph{mpc.cost.dx})
we use a parameterized version of the true dynamics.
In the pendulum domain, the parameters are the mass, length, and gravity;
and in the cartpole domain, the parameters are the cart's mass,
pole's mass, gravity, and length.
For cost learning in \emph{mpc.cost} and \emph{mpc.cost.dx} we
parameterize the cost of the controller as the weighted distance
to a goal state $C(\tau)=||w_g\circ(\tau-\tau_g)||_2^2$.
We have found that simultaneously learning the
weights $w_g$ and goal state $\tau_g$ is instable and
in our experiments we alternate learning of $w_g$
and $\tau_g$ independently every 10 epochs.
We collected a dataset of trajectories from an
expert controller and vary the number of trajectories our
models are trained on.
A single trial of our experiments takes 1-2 hours on a modern CPU.
We optimize the \emph{nn} setting with Adam \citep{kingma2014adam} with
a learning rate of $10^{-4}$ and all other settings are optimized
with RMSprop \citep{tieleman2012lecture} with a learning rate of
$10^{-2}$ and a decay term of $0.5$.

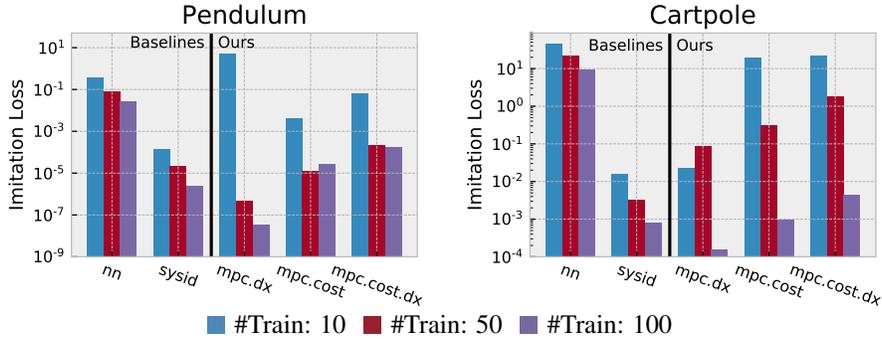
\begin{figure}[t]
  \centering
  \includegraphics[width=0.43\textwidth]{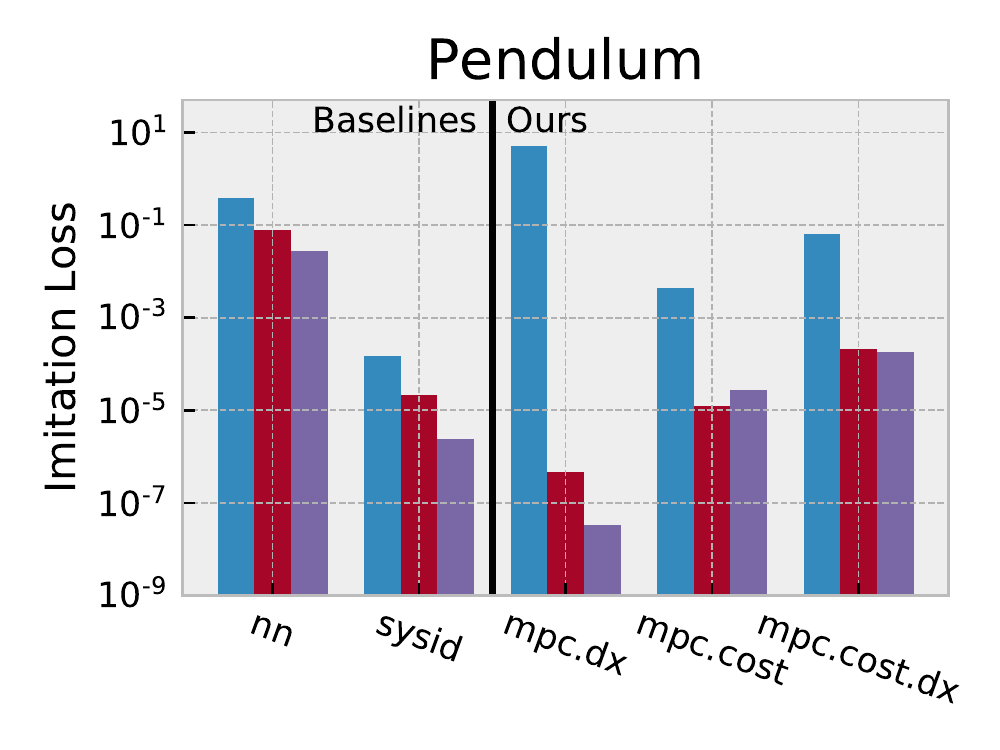}
  \includegraphics[width=0.43\textwidth]{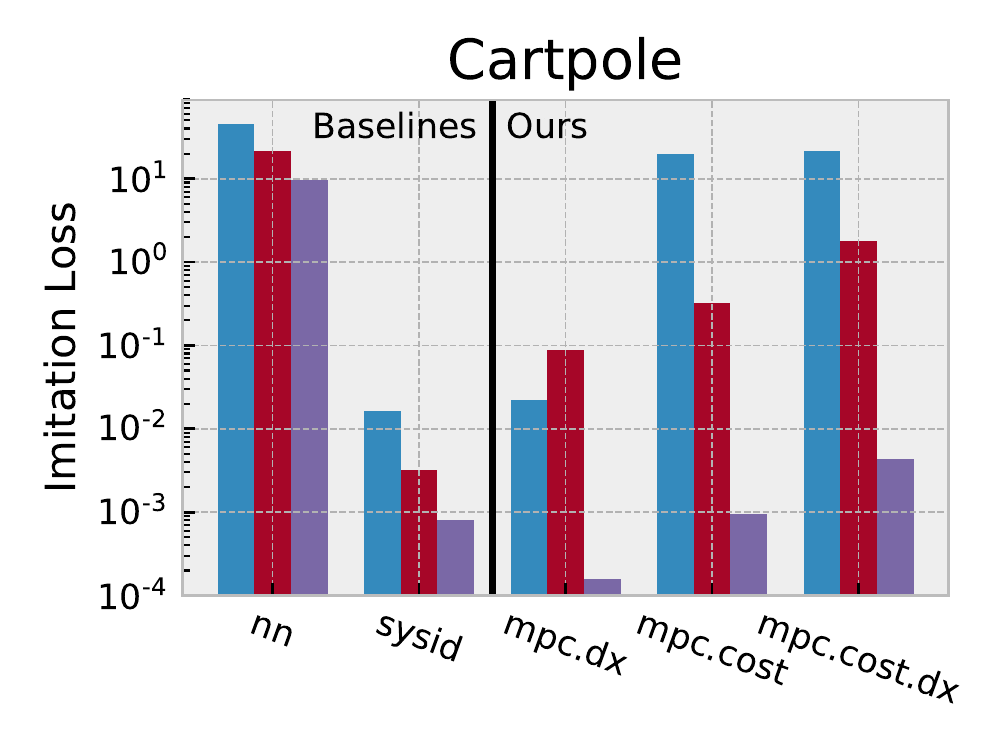}
  
  \vspace{-3mm}
  \begin{tikzpicture}
    [
    node/.style={square, minimum size=10mm, thick, line width=0pt},
    ]
    \node[fill={rgb,255:red,77;green,136;blue,184}] (n1) [] {};
    \node[] (n2) [right=0mm of n1] {\#Train: 10};
    \node[fill={rgb,255:red,152;green,32;blue,44}] (n3) [right=1mm of n2] {};
    \node[] (n4) [right=0mm of n3] {\#Train: 50};
    \node[fill={rgb,255:red,119;green,104;blue,162}] (n5) [right=1mm of n4] {};
    \node[] (n6) [right=0mm of n5] {\#Train: 100};
  \end{tikzpicture}

  \caption{
    Learning results on the (simple) pendulum and cartpole environments.
    We select the best validation loss observed during the training run
    and report the best test loss.
  }
  \label{fig:il:test}
  \vspace{-3ex}
\end{figure}

\Cref{fig:il:test} shows that in nearly every case we
are able to directly optimize the imitation loss with respect
to the controller and we significantly outperform a general
neural network policy trained on the same information.
In many cases we are able to recover the true cost function
and dynamics of the expert.
More information about the training and validation losses
are in \cref{sec:il-details}.
The comparison between our approach \emph{mpc.dx} and SysId is notable,
as we are able to recover equivalent performance to SysId
with our models using \emph{only} the control information and
\emph{without} using state information.

Again, while we emphasize that these are simple tasks, there are
stark differences between the approaches.
Unlike the generic network-based imitation learning, the MPC policy can
exploit its inherent structure.
Specifically, because the network contains a well-defined notion of the dynamics
and cost, it is able to learn with much lower sample complexity that a typical
network. But unlike pure system identification (which would be reasonable only
for the case where the physical parameters are unknown but all other costs are
known), the differentiable MPC policy can naturally be adapted to objectives
\emph{besides} simple state prediction, such as incorporating the additional
cost learning portion.

\newpage
\subsection{Imitation Learning: SysId with a non-realizable expert}
\label{il:non-realizable}

All of our previous experiments that involve SysId and learning
the dynamics are in the unrealistic case when the expert's dynamics
are in the model class being learned.
In this experiment we study a case where the expert's dynamics
are \emph{outside} of the model class being learned.
In this setting we will do imitation learning for the parameters of
a dynamics function with vanilla SysId and by directly optimizing
the imitation loss
(\emph{sysid} and the \emph{mpc.dx} in the previous section, respectively).

\begin{figure}[t]
  \centering
  \includegraphics[width=0.48\textwidth]{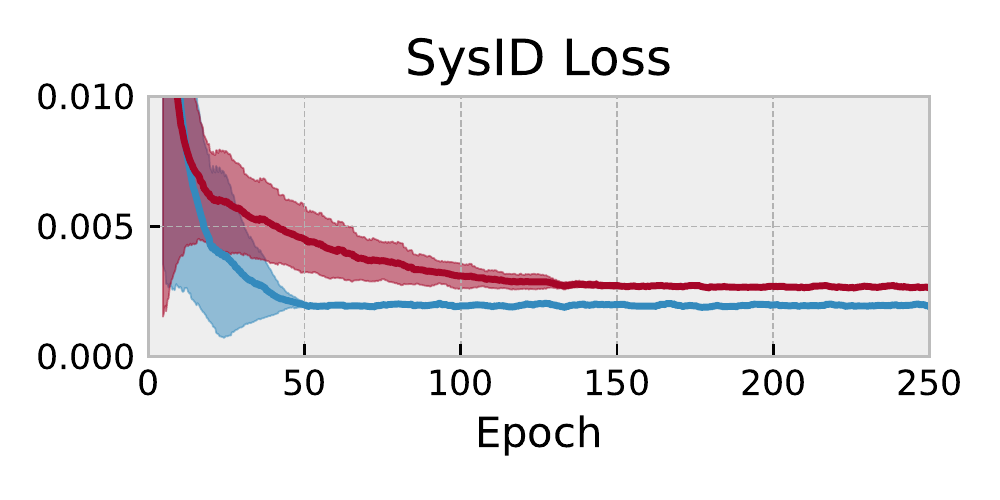}
  \includegraphics[width=0.48\textwidth]{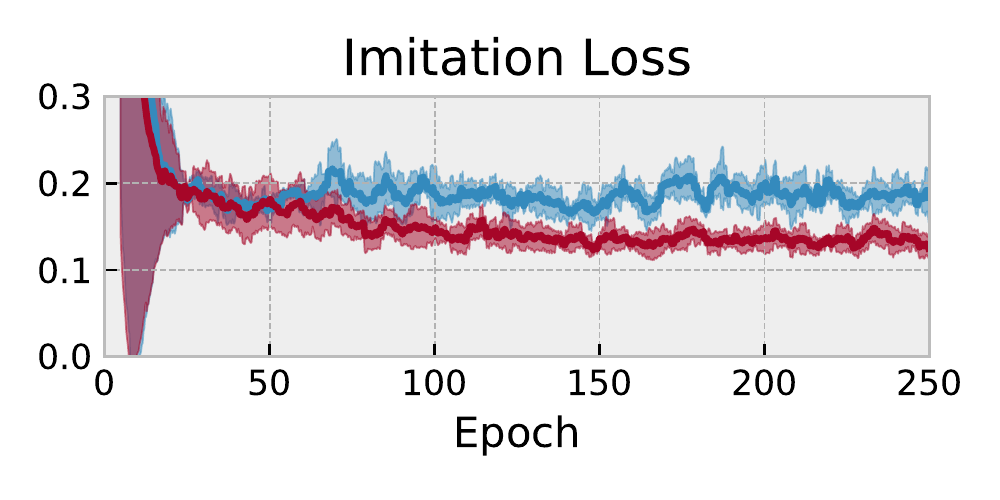}
  
  \vspace{-3mm}
  \begin{tikzpicture}
    [
    node/.style={square, minimum size=10mm, thick, line width=0pt},
    ]
    \node[fill={rgb,255:red,77;green,136;blue,184}] (n1) [] {};
    \node[] (n2) [right=0mm of n1] {Vanilla SysId Baseline};
    \node[fill={rgb,255:red,152;green,32;blue,44}] (n3) [right=1mm of n2] {};
    \node[] (n4) [right=0mm of n3] {(Ours) Directly optimizing the Imitation Loss};
  \end{tikzpicture}

  \caption{
    Convergence results in the non-realizable Pendulum task.
  }
  \label{fig:il:non-realizable}
  \vspace{-3ex}
\end{figure}
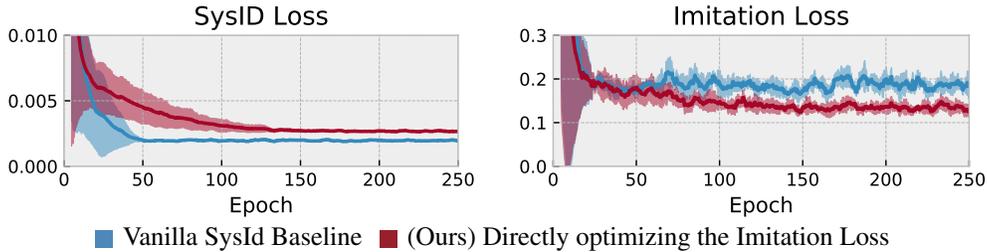

SysId often fits observations from a noisy environment to a simpler model.
In our setting, we collect optimal trajectories from an expert in
the pendulum environment that has an additional damping term and
also has another force acting on the point-mass at the end
(which can be interpreted as a ``wind'' force).
We do learning with dynamics models that \emph{do not} have these
additional terms and therefore we \emph{cannot} recover the
expert's parameters.
\Cref{fig:il:non-realizable} shows that
even though vanilla SysId is slightly better at optimizing the
next-state transitions, it finds an inferior model for imitation
compared to our approach that directly optimizes the imitation loss.

We argue that the goal of doing SysId is rarely in isolation and always
serves the purpose of performing a more sophisticated task such
as imitation or policy learning.
Typically SysId is merely a surrogate for optimizing the task
and we claim that the task's loss signal provides useful information
to guide the dynamics learning.
Our method provides one way of doing this by allowing the
task's loss function to be directly differentiated with respect to
the dynamics function being learned.

\newpage
\section{Conclusion}
This paper lays the foundations for differentiating and learning MPC-based
controllers within reinforcement learning and imitation learning. Our approach,
in contrast to the more traditional strategy of ``unrolling'' a policy,
has the benefit that it is much less computationally and memory intensive, with
a backward pass that is essentially free given the number of iterations required
for a the iLQR optimizer to converge to a fixed point. We have demonstrated our
approach in the context of imitation learning, and have highlighted the
potential advantages that the approach brings over generic
imitation learning and system identification.

We also emphasize that one of the primary contributions of this paper is
to define and set up the framework for differentiating through MPC in general.
Given the recent prominence of attempting to incorporate planning and
control methods into the loop of deep network architectures, the techniques here
offer a method for efficiently integrating MPC policies into such situations,
allowing these architectures to make use of a very powerful function class that
has proven extremely effective in practice.
The future applications of our differentiable MPC method include
tuning model parameters to task-specific goals and
incorporating joint model-based and policy-based loss functions;
and our method can also be extended for stochastic control.

\subsubsection*{Acknowledgments}
BA is supported by the National Science Foundation Graduate
Research Fellowship Program under Grant No.~DGE1252522.
We thank Alfredo Canziani, Shane Gu, Yuval Tassa,
and Chaoqi Wang for insightful discussions.

\bibliographystyle{plainnat}
{\small
\bibliography{empc}
}

\newpage
\appendix
\section{LQR and MPC Algorithms}
\label{sec:mpc-algs}

\begin{algorithm}[h]
\caption{
  $\mathrm{LQR}_T(\xinit; C, c, F, f)$ \hfill
  \emph{Solves \cref{eq:lqr} as described
  in \cite{levine2017optimal}}
}
The \textbf{state space} is $n$-dimensional
and the \textbf{control space} is $m$-dimensional. \\
$T\in\ZZ_+$ is the \textbf{horizon length}, the number of
nominal timesteps to optimize for in the future. \\
$\xinit\in\RR^n $ is the initial state \\
$C\in\RR^{T\times n+m\times n+m}$ and $c\in\RR^{T\times n+m}$ are the
quadratic cost terms. Every $C_t$ must be PSD. \\
$F\in\RR^{T\times n\times n+m}$ $f\in\RR^{T\times n}$ are the affine
cost terms.
\hrule
\label{alg:lqr}
\begin{algorithmic}
  \State \LeftComment \textbf{Backward Recursion}
  \State $V_T = v_T = 0$
  \For{t = T to 1}
  \State $Q_t = C_t+F_t^\top V_{t+1}F_t$
  \State $q_t = c_t+F_t^\top V_{t+1}f_t + F_t^\top v_{t+1}$
  \State $K_t = -Q_{t,uu}^{-1} Q_{t,ux}$
  \State $k_t = -Q_{t,uu}^{-1} q_{t,u}$
  \State $V_t = Q_{t,xx}+Q_{t,xu}K_t+K_t^\top Q_{t,ux}+K_t^\top Q_{t,uu}K_t$
  \State $v_t = q_{t,x}+Q_{t,xu}k_t+K_t^\top q_{t,u}+K_t^\top Q_{t,uu}k_t$
  \EndFor

  \State \LeftComment \textbf{Forward Recursion}
  \State $x_1 = \xinit$
  \For{t = 1 to T}
  \State $u_t = K_t x_t + k_t$
  \State $x_{t+1} = F_t \begin{bmatrix}x_t \\ u_t \end{bmatrix}+ f_t$
  \EndFor
  \State
  \State \Return $x_{1:T}, u_{1:T}$
\end{algorithmic}
\end{algorithm}

\newpage
\begin{algorithm}[H]
\caption{
  $\mathrm{MPC}_{T,\underline{u},\overline{u}}(\xinit, \uinit; C, f)$ \hfill
  \emph{Solves \cref{eq:ilqr} as described
  in \cite{tassa2014control}}
}
\label{alg:mpc}
The \textbf{state space} is $n$-dimensional
and the \textbf{control space} is $m$-dimensional. \\
$T\in\ZZ_+$ is the \textbf{horizon length}, the number of
nominal timesteps to optimize for in the future. \\
$\underline{u},\overline{u}\in\RR^m$ are respectively
  the control \textbf{lower-} and \textbf{upper-bounds}. \\
$\xinit\in\RR^n,\uinit\in\RR^{T\times m}$ are respectively the initial
  state and nominal control sequence \\
$C: \RR^{n\times m} \rightarrow \RR$ is the non-convex
  and twice-differentiable \textbf{cost function}. \\
$F: \RR^{n\times m} \rightarrow \RR^n$ is the non-convex
  and once-differentiable \textbf{dynamics function}.
\hrule
\begin{algorithmic}
  \State $x_1^1 = \xinit$
  \For {t = 1 to T-1}
  \State $x_{t+1}^1 = f(x_t, u_{{\rm init},t})$
  \EndFor
  \State $\tau^1 = [x^1, \uinit]$
  \State
  \For {i = 1 to \emph{[converged]}}
  \For {t = 1 to T}
    \State \(\triangleright\) Form the \emph{second-order Taylor expansion} of the
    cost as in \cref{eq:ilqr-cost-taylor}
    \State $C_t^i = \nabla_{\tau_t^i}^2 C(\tau_t^i)$
    \State $c_t^i = \nabla_{\tau_t^i} C(\tau_t^i) - (C_t^i)^\top  \tau_t^i$

    \State
    \State \(\triangleright\) Form the \emph{first-order Taylor expansion} of
    the dynamics as in \cref{eq:ilqr-dynamics-taylor}
    \State $F_t^i=\nabla_{\tau_t^i} f(\tau_t^i)$
    \State $f_t^i= f(\tau_t^i) - F_t^i \tau_t^i$
  \EndFor
  \State $\tau_{1:T}^{i+1} = \mathrm{MPCstep}_{T,\underline{u},\overline{u}}(
  \xinit, C, f, \tau_{1:T}^i, C^i, c^i, F^i, f^i)$
  \EndFor
  \State
  \Function{$\mathrm{MPCstep}_{T,\underline{u},\overline{u}}$}{
    $\xinit, C, f, \tau_{1:T}, \tilde C, \tilde c, \tilde F, \tilde f$}
  \State \(\triangleright\)
  $C,f$ are the \emph{true cost} and \emph{dynamics} functions.
  $\tau_{1:T}$ is the \emph{current trajectory} iterate.
  \State \(\triangleright\) $\tilde C, \tilde c, \tilde F, \tilde f$ are
  the \emph{approximate cost} and \emph{dynamics} terms around the current trajectory.
  \State
  \State \(\triangleright\) \textbf{Backward Recursion:} Over the linearized trajectory.
  \State $V_T = v_T = 0$
  \For{t = T to 1}
  \State $Q_t = \tilde C_t+\tilde F_t^\top V_{t+1}\tilde F_t$
  \State $q_t = \tilde c_t+\tilde F_t^\top V_{t+1}\tilde f_t + \tilde F_t^\top v_{t+1}$
  \State
  \State $k_t = \argmin_{\delta u} \;\;
        \frac{1}{2} \delta u^\top  Q_{t,uu}\delta u + Q_x^\top  \delta u
    \;\; \mathrm{s.t.} \;\; \underline{u}\leq u+\delta u \leq \overline{u}$
  \State \(\triangleright\) Can be solved with a \emph{Projected-Newton
  method} as described in \cite{tassa2014control}.
  \State \(\triangleright\) Let $f,c$ respectively index the \emph{free} and
   \emph{clamped} dimensions of this optimization problem.
  \State
  \State $K_{t,f} = -Q_{t,uu,f}^{-1} Q_{t,ux}$
  \State $K_{t,c} = 0$
  \State
  \State $V_t = Q_{t,xx}+Q_{t,xu}K_t+K_t^\top Q_{t,ux}+K_t^\top Q_{t,uu}K_t$
  \State $v_t = q_{t,x}+Q_{t,xu}k_t+K_t^\top q_{t,u}+K_t^\top Q_{t,uu}k_t$
  \EndFor

  \State
  \State \(\triangleright\) \textbf{Forward Recursion and Line Search:}
  Over the true cost and dynamics.
  \Repeat
  \State $\hat x_1 = \tau_{x_1}$
  \For{t = 1 to T}
  \State $\hat u_t = \tau_{u_t} + \alpha k_t +  K_t (\hat x_t - \tau_{x_t})$
  \State $\hat x_{t+1} = f(\hat x_t, \hat u_t)$
  \EndFor
  \State $\alpha = \gamma \alpha$
  \Until {$\sum_t C([\hat x_t, \hat u_t]) \leq \sum_t C(\tau_t)$}
  \State
  \State \Return $\hat x_{1:T}, \hat u_{1:T}$
  \EndFunction
\end{algorithmic}
\end{algorithm}

\newpage
\section{Imitation learning experiment losses}
\label{sec:il-details}

\begin{figure}[h]
  \centering
  \includegraphics[width=\textwidth]{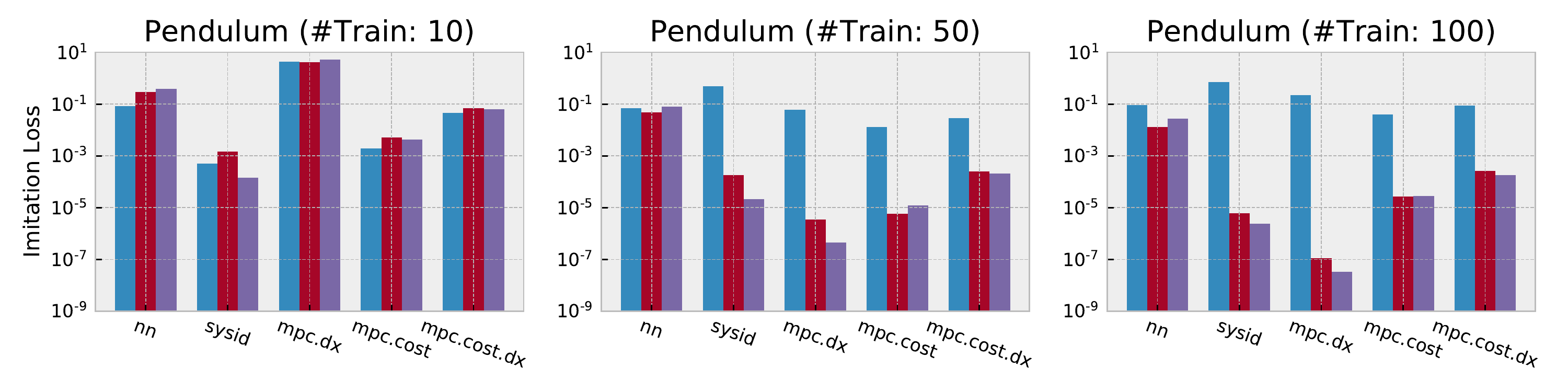} \\
  \includegraphics[width=\textwidth]{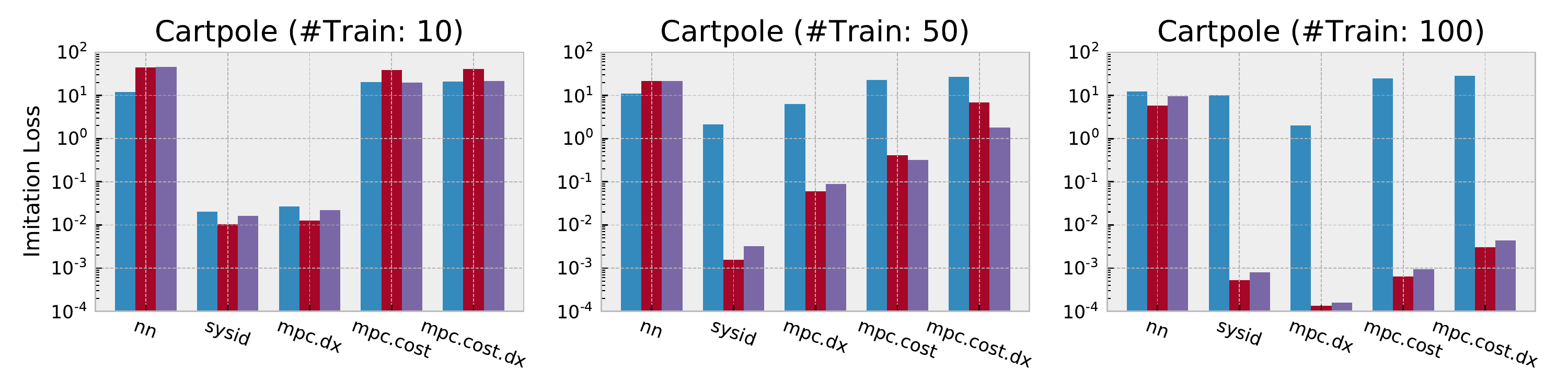}
  
  \begin{tikzpicture}
    [
    node/.style={square, minimum size=10mm, thick, line width=0pt},
    ]
    \node[fill={rgb,255:red,77;green,136;blue,184}] (n1) [] {};
    \node[] (n2) [right=0mm of n1] {Train};
    \node[fill={rgb,255:red,152;green,32;blue,44}] (n3) [right=1mm of n2] {};
    \node[] (n4) [right=0mm of n3] {Val};
    \node[fill={rgb,255:red,119;green,104;blue,162}] (n5) [right=1mm of n4] {};
    \node[] (n6) [right=0mm of n5] {Test};
  \end{tikzpicture}

  \caption{
    Learning results on the (simple) pendulum and cartpole environments.
    We select the best validation loss observed during the training run
    and report the corresponding train and test loss.
    Every datapoint is averaged over four trials.
  }
  \label{fig:il:all}
  \vspace{-3ex}
\end{figure}
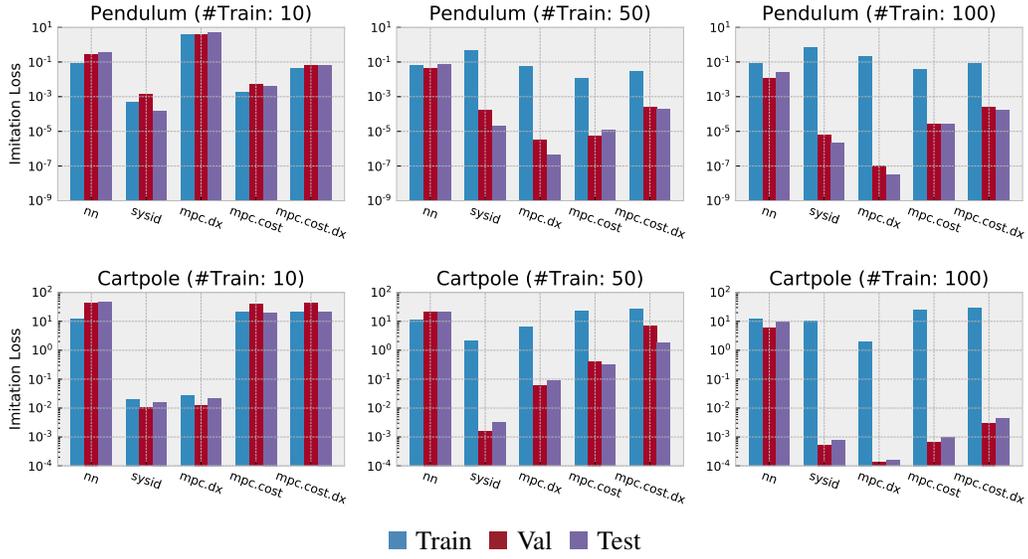

\end{document}